\documentclass[letterpaper]{article} 
\usepackage{aaai25}  
\usepackage{times}  
\usepackage{helvet}  
\usepackage{courier}  
\usepackage[hyphens]{url}  
\usepackage{graphicx} 
\usepackage{natbib}  
\usepackage{caption} 
\usepackage{algorithm}
\usepackage{newfloat}
\usepackage{listings}
\DeclareCaptionStyle{ruled}{labelfont=normalfont,labelsep=colon,strut=off} 
\floatstyle{ruled}
\newfloat{listing}{tb}{lst}{}
\floatname{listing}{Listing}
\usepackage[switch, modulo]{lineno}

\title{Decoupling Generation and Evaluation for Parallel Greedy Best-First Search(extended version)}
\author{
    Takumi Shimoda\textsuperscript{},
    Alex Fukunaga\\
}
\affiliations{
    
    Graduate School of Arts and Sciences\\
    The University of Tokyo\\
    takumi35shimoda@yahoo.co.jp, fukunaga@idea.c.u-tokyo.ac.jp
}

\usepackage{comment}
\usepackage{xspace}
\usepackage{bm}

\usepackage{subcaption}
\usepackage{multirow}
\usepackage{algorithm}
\usepackage{amsmath}
\usepackage{amsfonts}
\usepackage{amsthm}
\usepackage[noend]{algpseudocode}
\usepackage{booktabs} 
\usepackage{xr} 
\usepackage{amssymb} 
\usepackage{varwidth}
\theoremstyle{definition}
\newtheorem{def.}{Definition}
\newtheorem{th.}{Theorem}

\newtheorem{lem.}{Lemma}
\newtheorem{cor.}{Corollary}
\newtheorem{prop.}{Proposition}
\newtheorem{obs.}{Observation}

\newcommand{\pddl}[1]{\textsf{\small #1}}

\newcommand{\open}{\mathit{Open}}
\newcommand{\closed}{\mathit{Closed}}

\newcommand{\true}{\mathit{true}}
\newcommand{\false}{\mathit{false}}

\newcommand{\satisfies}{\mathit{satisfies}}

\newcommand{\NULL}{\mathit{NULL}}
\newcommand{\Succ}{\mathit{succ}}
\newcommand{\hwm}{\mathit{hwm}}

\newcommand{\BTS}{\mathit{BTS}}

\newif\ifwastar
\wastarfalse

\newcommand{\Unevaluated}{\mathit{Unevaluated}}   

\newcommand{\PDUHF}{PUHF3$_S$\xspace}
\newcommand{\KPDGBFS}{KPGBFS$_S$\xspace}
\newcommand{\OBATS}{OBAT$_S$\xspace}

\algblock{Indent}{EndIndent}
\algnotext{Indent} 
\algnotext{EndIndent} 
\begin{document}

\maketitle

\begin{abstract}
  In order to understand and control the search behavior of parallel search, recent work has proposed a class of constrained parallel greedy best-first search algorithms which only expands states that satisfy some constraint.
  However, enforcing such constraints can be costly, as threads must be waiting idly until a state that satisfies the expansion constraint is available.
  We propose an improvement to constrained parallel search which decouples state generation and state evaluation and significantly improves state evaluation rate, resulting in better search performance.
\end{abstract}

\section{Introduction}

Parallelization of combinatorial search algorithms is important
in order to maximize search algorithm performance on
modern, multi-core CPUs.
Greedy Best First Search (GBFS), which repeatedly expands the best state from $\open$ according to an
evaluation function 
and a tie-breaking strategy, is 
a widely used satisficing search algorithm \cite{DoranM66}.
However, the performance of straightforward parallelizations of GBFS is non-monotonic 
-- there is a significant risk that using $k$ threads can result in significantly worse performance than using fewer than $k$ threads.
It has been shown experimentally that parallel GBFS can expand orders of magnitude more states than GBFS \cite{KuroiwaF2019},  and it has been shown theoretically that parallel GBFS using a shared $\open$ and/or $\closed$ list (including KPGBFS, a straightforward parallelization of GBFS) can expand arbitrarily many more states than GBFS \cite{KuroiwaF2020}.

Recently, constrained parallel GBFS algorithms which are guaranteed to only expand states which satisfy some {\it expansion constraint} have been proposed. 
PUHF \cite{KuroiwaF2020} constrains the search so that only states in 
the {\it Bench Transition System} ($\BTS$), the set of all states that can be expanded by GBFS under some tie-breaking order \cite{HeusnerKH17} are expanded. 
OBAT \cite{ShimodaF25} further constrains the search so that multiple benches are not simultaneously explored.

However, such constraints incur a cost, 
as threads can be forced to be idle while waiting for a 
state which is guaranteed to satisfy the expansion constraint becomes available.
Due to this idle waiting (poor CPU utilization), it has been shown that PUHF and OBAT both have a significantly lower state expansion rate compared to unconstrained parallel GBFS. As a result, even on problems where constrained parallel GBFS finds a solution with fewer expansions  than unconstrained parallel GBFS, constrained parallel GBFS can underperform unconstrained parallel GBFS.

We propose Separate Generation and Evaluation (SGE), which decouples state expansion and evaluation so that instead of waiting for a single thread to fully expand a state (generating and evaluating its successors), multiple threads evaluate the successors.
We show that this significantly improves the state evaluation rate in constrained parallel GBFS, resulting in significantly improved performance.

\section{Preliminaries and Background}
\label{sec:preliminaries}

\subsubsection {K-Parallel GBFS (KPGBFS)}

K-Parallel BFS \cite{VidalBH10} is a straightforward, baseline parallelization of BFS.
All threads share a single $\open$ and $\closed$.
Each thread locks $\open$ to remove
a state $s$ with the lowest $f$-value in $\open$, locks $\closed$ to
check duplicates and add $\Succ(s)$ to $\closed$, and locks $\open$ to
add $\Succ(s)$ to Open.
KPGBFS is KPBFS with $f(s)=h(s)$.

\subsubsection{Constrained Parallel GBFS}
\label{sec:cpgbfs}

Recent work has investigated parallel GBFS algorithms based on KPGBFS which expand states only if they satisfy some expansion constraint.
Parallel Under High-water mark First (PUHF) and its successors PUHF2--4 \cite{KuroiwaF2020,ShimodaF23} expanded only states which are guaranteed to be in the $\BTS$ \cite{HeusnerKH17}, but unlike sequential GBFS, PUHF can simultaneously expand states in multiple benches.
One Bench At a Time (OBAT) further constrains the search to expand states in a single bench at a time, guaranteeing that the number of states expanded is bounded relative to sequential GBFS with some time-breaking order \cite{ShimodaF25}.

Constrained Parallel GBFS (CPGBFS) (Algorithm \ref{alg:cpgbfs}) is a schema  for a class of parallel search algorithms based on KPGBFS, which 
only expands nodes which satisfy some algorithm-specific constraint in line \ref{cpgbfs:expansion-constraint}, 
where 
$\satisfies$ is a function which returns $\true$ if and only if  $s$ satisfies the algorithm-specific expansion constraint.

KPGBFS is a special case of CPGBFS where $\satisfies$ always returns $\true$.
The previously proposed constrained search algorithms (PUHF, PUHF2--4, and OBAT) are instances of CPGBFS where 
the $\satisfies$ function implements a check for the sufficient constraint which guarantees that $s$ is in the BTS (and in the case of OBAT, further constrains the expansion to prevent simultaneous expansion in multiple benches) -- the specific implementation details of $\satisfies$ depend on the specific algorithm. 
\footnote{The original presentations of PUHF presented these algorithms as marking states guaranteed to be in the BTS as {\it certain}, and only expanding nodes marked as {\it certain}, but it is straightforward to reframe this as a constraint check as in Algorithm \ref{alg:cpgbfs}.}

\begin{algorithm}[tb]
  \begin{algorithmic}[1]
    \footnotesize
    \State $\open \leftarrow \{ s_\textit{init} \}, \closed \leftarrow \{ s_\textit{init} \}; \forall i, s_i \leftarrow \NULL$
    \For{$i \leftarrow 0,...,k - 1$ in parallel}
      \Loop
        \State \textbf{with} lock($\open$)
        \Indent
        \If{$\open = \emptyset$} 
           \If{$\forall j, s_j = \NULL$}
           \State{unlock($\open$); \Return $NULL$} \EndIf
        \ElsIf{$\satisfies(top(\open)) = \true$}       \label{cpgbfs:expansion-constraint} 
          \State $s_i \leftarrow top(\open)$; $\open \leftarrow \open \setminus \{ s_i \}$
        \EndIf
        \EndIndent
        \If{$s_i = \NULL$} continue \EndIf
        \If{$s_i \in s_\textit{goal}$} \Return $Path(s_i)$ \EndIf
        \For{$s_i' \in \Succ(s_i)$}
          \State lock($\closed$)
          \If{$s_i' \notin \closed$}
            \State $\closed \leftarrow \closed \cup \{ s_i' \}$ 
            \State unlock($\closed$)
            \State $children(s_i) \leftarrow children(s_i) \cup \{ s_i' \}$ 
            \State evaluate($s'$)
          \Else
            \State unlock($\closed$)
          \EndIf
        \EndFor
        \State \textbf{with} lock($\open$) \label{cpgbfs:begin-insert-successors}
        \Indent
        \For{$s_i' \in children(s_i)$}
          \State  $\open \leftarrow \open \cup \{ s_i' \}$
        \EndFor \label{cpgbfs:end-insert-successors}
        \EndIndent
        \State $s_i \leftarrow \NULL$
      \EndLoop
    \EndFor
        \caption{CPGBFS: Constrained Parallel GBFS}
    \label{alg:cpgbfs}
  \end{algorithmic}
\end{algorithm}

\section{State Expansion Bottlenecks in Constrained Parallel Search}
\label{sec:bottleneck}

All previous CPGBFS variants (all PUHF variants and OBAT) have a significantly lower state evaluation rate than unconstrained parallel search (KPGBFS).
There are two related reasons: (1) the expansion constraint, and (2) batch successor insertion.

\subsubsection{Expansion Constraint Bottleneck}
\begin{figure}[tb]
  \centering
  \includegraphics[width=1.0\columnwidth,,clip]{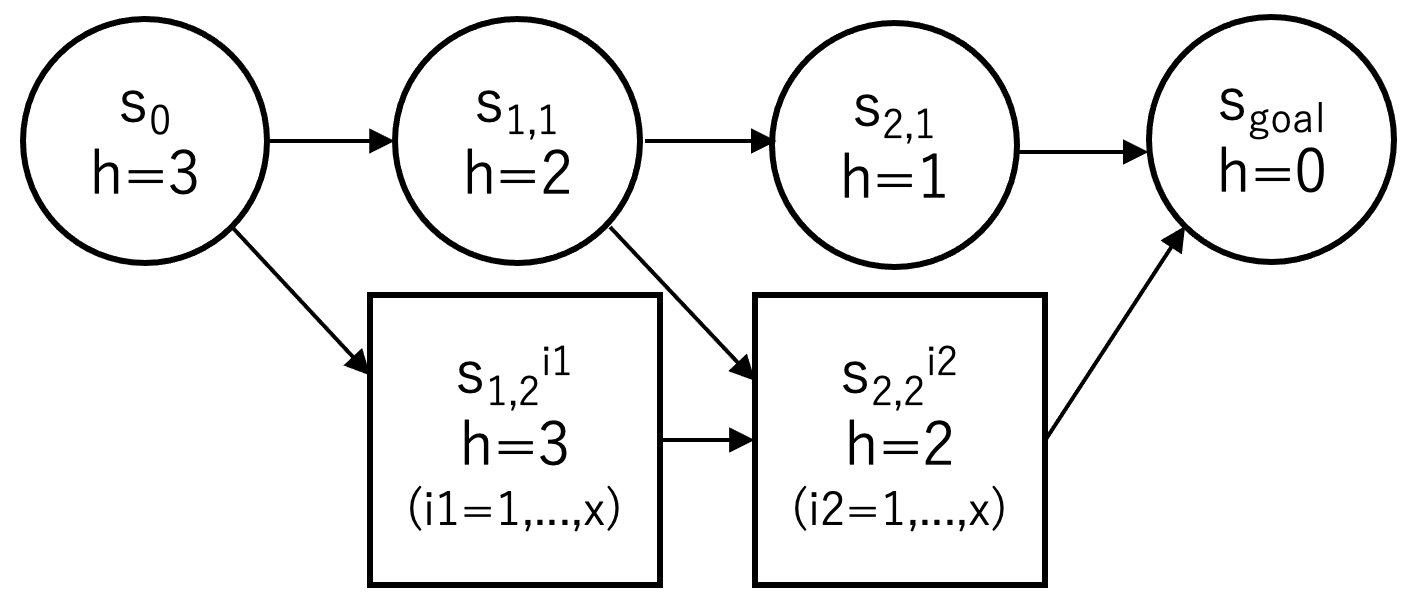}
  \caption{Example of parallel expansion bottleneck}
  \label{fig:divided-image}
\end{figure}

Unconstrained parallel search algorithms such as KPGBFS will unconditionally expand the top states in $\open$.
Threads in unconstrained algorithms are only idle when waiting for a mutex lock for the shared $\open$ and $\closed$ structures, and  
such mutex overheads can be greatly reduced by efficiently implementing $\open$/$\closed$ (e.g., sharded implementations).

In contrast, CPGBFS cannot expand the top state $top(\open)$ unless the expansion constraint is satisfied, 
even if the mutex lock is available.
For example, in Figure~\ref{fig:divided-image} the only states in the $\BTS$ are the circular nodes ($s_0,s_{1,1},s_{2,1},s_\textit{goal}$), so any algorithm which constrains expansions to states in the $\BTS$ can only expand one of these states at a time, while all other threads must wait.


\subsubsection{Batch Successor Insertion Bottleneck}

Consider Figure~\ref{fig:divided-image}. In a standard, single-thread implementation of best-first search with eager evaluation, the expansion of $s_{1,1}$ includes (1) generating $succ(s_{1,1})=s_{2,1},s_{2,2}^1,\ldots,s_{2,2}^x$, (2) evaluating all states in
$succ(s_{1,1})$ with a heuristic evaluation function, and (3) inserting $succ(s_{1,1})$ in $\open$.
In many cases, computing the heuristic evaluation function consumes the majority of time spent expanding the state, and 
the full expansion of a single state such as $s_{1,1}$ can take a significant amount of time due to the evaluation of all of its successors.

A constrained parallel search which seeks to expand a similar set of nodes as GBFS has an additional requirement not present in single-threaded search: the successors of a state $s$ are all simultaneously inserted in $\open$ only after all successors of state $s$ are evaluated (Algorithm \ref{alg:cpgbfs}, lines \ref{cpgbfs:begin-insert-successors}--\ref{cpgbfs:end-insert-successors}).
This batch insertion ensures that the successors of $s$ are expanded in best-first order -- otherwise (e.g., if states were inserted one at a time directly into $\open$ immediately after being evaluated),  a state $s'$ with a worse $f$-value than its sibling $s''$ might be inserted into $\open$ and expanded by another thread before $s''$ has been inserted into $\open$.

The state evaluation rate of unconstrained parallel search (e.g., KPGBFS), is not significantly affected by whether the successors are inserted in a single batch or one at a time, because available threads can freely expand the top states from $\open$.
However, in CPGBFS, the combination/interaction of the expansion constraint and the batch successor insertion requirement causes a significant bottleneck.
For example,  in Figure~\ref{fig:divided-image}, in Algorithm \ref{alg:cpgbfs}, if a thread starts to expand $s_{1,1}$, all other threads must stop and wait until all of $succ(s_{1,1})$ have been fully evaluated and inserted into $\open$.

\section{Separate Generation and Evaluation}
\label{sec:divided}

We propose Separate Generation and Evaluation (SGE), an approach for increasing the state evaluation rate in CPGBFS.
SGE alleviates the batch successor insertion bottleneck described above.

Continuing the Figure~\ref{fig:divided-image} example from the previous section, 
instead of waiting idly while one thread expands $s_{1,1}$ (which includes computing all of the heuristic values for $s_{2,1},s_{2,2}^1,..,s_{2,2}^x$), 
it would be more efficient to parallelize the evaluation of $s_{2,1},s_{2,2}^1,..,s_{2,2}^x$ among available threads.

\begin{algorithm}[tb]
  \begin{algorithmic}[1]
      \footnotesize
      \State {${\open \leftarrow\{s_\textit{init}\}, \closed\leftarrow\{s_\textit{init}\}; \forall{i}, s_i\leftarrow{\NULL}}$}
      \For{$i \leftarrow {0, ..., k-1}$ in parallel}
          \Loop
              \State lock($\Unevaluated$)
              \If{$\Unevaluated\neq\emptyset$} \label{cpgbfs-sge:unevaluated-notempty}
                  \State{$ s_i \gets top(Unevaluated)$}
                  \State{$\Unevaluated \leftarrow \Unevaluated  \setminus \{s_i\}$}
                  \State{unlock($\Unevaluated$)}
                  \State{evaluate($s_i$)} \Comment{cached evaluation}
                  \State{\textbf{with} lock($Evaluated\:Successor$)}
                  \Indent
                  \State{$Evaluated\:Successor($parent of $\{s_i\}) \leftarrow $}
                  \State{$\:\:Evaluated\:Successor($parent of $\{s_i\}) \cup \{s_i\}$}
                  \If{all siblings of $s_i$ have been evaluated} \label{cpgbfs-sge:insert-open-start}
                      \State{\textbf{with} lock($\open$), lock($\closed$)}
                      \Indent
                      \For{$s_i' \in  $ siblings of $s_i$}
                          \If{$s_i' \notin \closed$}
                            \State $\closed \leftarrow {\closed \cup \{s_i'\}}$
                            \State {$\open\gets \open\cup\{s_i'\}$}
                          \EndIf
                      \EndFor
                      \EndIndent
                  \EndIf \label{cpgbfs-sge:insert-open-end}
                  \EndIndent
              \Else
                  \State {unlock($\Unevaluated$)}
                  \State {\textbf{with} lock($\open$)}
                  \Indent
                  \If{$\open = \emptyset$}
                      \State {unlock($\open$)}
                      \If {$\forall{j},s_j=\NULL$}    
                          \State {\Return $\NULL$}
                      \EndIf
                  \ElsIf{$\satisfies(top(\open)) = \true$}
                        \State $s_i \leftarrow top(\open)$; $\open \leftarrow \open \setminus \{ s_i \}$
                  \EndIf
                  \EndIndent
                  \If{$s_i = \NULL$} continue \EndIf
                  \If {$s_i \in s_\textit{goal}$} \Return $Path(s_i)$ \EndIf
                  \State \textbf{with} lock($\Unevaluated$)
                  \Indent
                  \State{$\Unevaluated\gets{\Unevaluated\cup succ(s_i')}$}
                  \EndIndent
              \EndIf        
              \State $s_i \leftarrow{\NULL}$
          \EndLoop
      \EndFor
      \caption{Constrained Parallel GBFS with SGE}
      \label{alg:cpgbfs-sge}   
  \end{algorithmic}
\end{algorithm}

The main idea of SGE is to decompose the expansion of state $s$ into separate units of work which can be parallelized: 
(1) successor generation, which generates $succ(s)$, the successors of $s$, and
    (2) successor evaluation, which evaluates $succ(s)$.
Algorithm~\ref{alg:cpgbfs-sge} shows Constrained Parallel GBFS with SGE.
    After a thread selects $s$ for expansion from the shared $\open$, it generates $succ(s)$, and inserts $succ(s)$  into the shared $\Unevaluated$ queue.
    The evaluation of states in $\Unevaluated$ is done in parallel, taking precedence over selection of states for expansion (a thread will select a state for expansion from $\open$ only if $\Unevaluated$ is currently empty (Algorithm \ref{alg:cpgbfs-sge}, line \ref{cpgbfs-sge:unevaluated-notempty})).

    Evaluated states are {\it not} immediately inserted into $\open$. Instead, we insert all of the successors of $s$ simultaneously into $\open$, after they have all been evaluated (lines \ref{cpgbfs-sge:insert-open-start}--\ref{cpgbfs-sge:insert-open-end}).  This is so that the parallel search is able to prioritize $succ(s)$ similarly to GBFS.

Consider the behavior of  PUHF (which only expands states in the $\BTS$) with SGE on the search space in Figure~\ref{fig:divided-image}.
First, a thread pops $s_0$ from $\open$ ($\satisfies(s_0)=\true$), and generates its successors ($s_{1,1},s_{1,2}^1,..,s_{1,2}^x$), which are inserted in $\Unevaluated$.
Available threads will pop these successors from $\Unevaluated$ and evaluate them.
When all successors of $s_0$ have been evaluated, they are all inserted in $\open$.
Next, some thread removes $s_{1,1}$ ($\satisfies(s_{1,1})=\true$), generates its successors ($s_{2,1},s_{2,2}^1,..,s_{2,2}^x$), and inserts them in $\Unevaluated$.
While generating the successors of $s_{1,1}$,  other threads may try to pop a state from $\open$, but since the top state at that time ($s_{1,2}^i$) is not in the $\BTS$ ($\satisfies(s_{1,2}^i) = \false$), $\open$ will remain untouched.
After the successors of $s_{1,1}$ have been inserted in $\Unevaluated$, the available threads will remove them from $\Unevaluated$ and evaluate them.
The search continues until $s_\textit{goal}$ is found.
Each time a state's successors are generated, available threads evaluate the successors in parallel. This clearly improves thread utilization compared to CPGBFS without SGE,
where only 1 thread is active throughout the search space in Figure \ref{fig:divided-image}. 

The basic idea of decoupling state generation and state evaluation is
similar to that of ePA*SE, which decouples state generation and {\it edge evaluations} in a parallel A* \cite{MukherjeeAL22}, where an edge evaluation is the computation required to evaluate the application of an operator, (e.g., collision checking using a simulation model in robot motion planning).
Because the requirements and objectives of GBFS (satisficing search) and A* (cost-optimal search) differ, the implementation of SGE is somewhat simpler (using an $\Unevaluated$ queue instead of dummy/real edges as in ePA*SE).

\subsubsection{SGE and search behavior}

The state expansion order of a parallel search algorithm $A$ with SGE will differ from the expansion order of $A$ without SGE.
Although it is nontrivial to precisely characterize the difference between the expansion order of an algorithm with and without SGE,
a simple approximation is that executing a parallel search algorithm $A$ with SGE on $k$ threads is somewhat similar to executing $A$ without SGE on $m < k$ threads, where each of the $m$ ``threads'' is faster than each of the actual $k$ threads.

As a simple example, consider searching a state space which is a tree with uniform branching factor 2, where the heuristic evaluation function computation is the computational bottleneck, and assume that $\open$ currently contains many nodes.
With $k=16$ threads, KPGBFS will expand 16 states at a time -- each thread expands 1 state, where the expansion includes generation and heuristic evaluation of the state's 2 successors.
In contrast, KPGBFS with SGE would be evaluating 16 states at a time -- each thread, after quickly generating the successors of a state, would then be assigned to evaluate 1 successor state, essentially the same as KPGBFS without SGE expanding 8 states, i.e., similar to KPGBFS without SGE running on 8 threads.

The efficiency (number of states expanded) of parallel GBFS compared to sequential GBFS tends to worsen as the number of threads increases,
so the tendency of SGE to cause the parallel search to behave as if there were fewer threads can result in more efficient search (fewer state expansions), compared to parallel search without SGE.

\subsubsection{SGE and expansion constraints}
The state expansion constraints (i.e., the $\satisfies$ check) for the various CPGBFS algorithms known to date are defined based on:
(a) comparisons between the $h$-value of a state's parent and the $h$-values of the siblings of $s$  (for PUHF), or 
(b) $h(s)$ vs. the $h$-values of other states currently being expanded (for PUHF2--4, OBAT).
Therefore, distributing the evaluation of the successors of $s$ among multiple threads has no impact on the correctness of the expansion constraint (i.e., the guarantee that the node being expanded is in the BTS). 

\section{Experimental Evaluation of SGE}
\label{sec:divided-experiment}


We evaluated SGE using the planning benchmark set used to evaluate parallel GBFS variants in \cite{ShimodaF25}.
These are based on the Autoscale-21.11 benchmark set (42 STRIPS domains, 30 instances/domain, 1260 total instances) \cite{torralba-et-al-icaps2021}, except that \pddl{gripper} and \pddl{miconic} were replaced with harder instances because the original Autoscale instances were too easy to distinguish among the parallel algorithms.
All search algorithms use the FF heuristic \cite{Hoffmann01}. 
Each run has a run time limit of 5 minutes and 3 GB RAM/thread (e.g., 24 GB total for a 8-thread run) limit on a Intel(R) Xeon(R) CPU E5-2670 v3 @ 2.30GHz processor.

We evaluated KPGBFS, KPGBFS with SGE (\KPDGBFS), PUHF3, PUHF3 with SGE (\PDUHF), OBAT, and OBAT with SGE (\OBATS) on $k \in \{4, 8, 16\}$ threads.
We also report baseline single-threaded GBFS results.
All tie-breaking is First-In-First-Out.
The code is available at \url{https://github.com/TakuShimoda/AAAI25}.

Table~\ref{tab:evaluation_rate} 
compares the state evaluation rates of the algorithms.
Table~\ref{tab:coverage} shows the number of instances solved by each algorithm.
In Table~\ref{tab:evaluation_rate}, as well as \% improvements mentioned below, we include only the 354
instances solved by all algorithms so that means can be computed.
Scatterplots of evaluation rate, number of states expanded, and search time are in Supplement.

\begin{table}[t]
  \begin{small}
  \begin{subtable}[h]{0.48\textwidth}
  \centering
  \begin{tabular}{|l|r|r|r|r|}
    \hline
    \#threads &  1 thread & 4 threads & 8 threads & 16 threads \\
    \hline
    GBFS  & 4814 & \multicolumn{3}{|c|}{-} \\    
    \hline
    KPGBFS  &-  &15271 & 27521 & 48772\\
    KPGBFS$_S$ & - & 15030 & 26309&  45668 \\
    \hline
    PUHF3 & - &  11804& 16806 & 22121 \\
    PUHF3$_S$ & - &  13295 &  21156 &  31814 \\
    \hline
    OBAT & - & 10221 &12830 & 15659 \\
    OBAT$_S$ & - & 12407 & 18713 & 24634 \\
    \hline    
  \end{tabular}
  \caption{State evaluation rate (states/second, geometric mean)}
  \label{tab:evaluation_rate}
  \end{subtable}
  \hfill

  \begin{subtable}[h]{0.48\textwidth}
  \centering
  \begin{tabular}{|l|r|r|r|r|}
    \hline
    \#threads &  1 thread & 4 threads & 8 threads & 16 threads \\
    \hline
    GBFS  & 401 & \multicolumn{3}{|c|}{-} \\
    \hline
    KPGBFS  &-  &462 & 488 & 529\\
    KPGBFS$_S$ & - & 472 & 500 & 532 \\
    \hline
    PUHF3 & - & 459 & 477 & 494 \\
    PUHF3$_S$ & - & 468 & 494 & 510 \\    
    \hline
    OBAT & - & 458 & 477 & 496 \\
    OBAT$_S$ & - & 478 & 506 & 532 \\ 
    \hline
  \end{tabular}
  \caption{Coverage results (out of 1260 total instances)}
  \label{tab:coverage}  
  \end{subtable}
  
  \end{small}  
  \caption{Autoscale-21.11/IPC-based benchmark results (1260 instances total). Means in
    Table~\ref{tab:evaluation_rate} 
    are for 354 instances solved by all algorithms}  

\end{table}

\subsubsection{Evaluation Rates}

Table~\ref{tab:evaluation_rate} shows that the constrained algorithms (PUHF3, OBAT) have a significantly lower evaluation rate (states/second) than the unconstrained KPGBFS.
The evaluation rate of \KPDGBFS is somewhat lower than KPGBFS for $k \in \{4,8,16\}$ threads. Thus, management of the overhead of a separate $\Unevaluated$ queue in  SGE imposes a noticeable evaluation rate penalty for an unconstrained search.
The effect of this overhead is noticeable for the problems with the highest evaluation rate (Fig~\ref{sge-supp:fig:evaluation-rate-comparisons} in Supplement). 
On the other hand, \PDUHF and \OBATS have significantly higher evaluation rates than PUHF3 and OBAT, respectively, showing that SGE successfully achieves the goal of improving the evaluation rate for constrained parallel best-first search.



\subsubsection{Number of States Expanded}

For 16 threads, \KPDGBFS, \PDUHF, \OBATS expanded 17.4\%, 14.8\%, 10.5\% fewer states than KPGBFS, PUHF3, OBAT, respectively.
Thus, SGE has the effect of reducing  the search required to solve problem instances for both constrained and unconstrained parallel GBFS.
See Fig~\ref{sge-supp:fig:expansions-comparisons} in Supplement for details. 

\subsubsection{Search Time}

The differences in state evaluation rate and search efficiency result in significantly improved search times overall.
For 16 threads, \KPDGBFS, \PDUHF, \OBATS, had 14.1\%, 39.1\%, 38.5\% faster search time than KPGBFS, PUHF3 and OBAT, respectively.
For 8 threads, \KPDGBFS, \PDUHF, \OBATS, had 12.4\%, 27.5\%, 27.8\% faster search time than KPGBFS, PUHF3 and OBAT, respectively.
For 4 threads, \KPDGBFS, \PDUHF, \OBATS, had 11.2\%, 19.7\%, 17.6\% faster search time than KPGBFS, PUHF3 and OBAT, respectively.
See Fig~\ref{sge-supp:fig:search-time-comparisons} in Supplement for details. 


\subsubsection{Coverage}

Table \ref{tab:coverage} shows that
SGE significantly improves the number of instances solved by PUHF3 and OBAT.

\section{Discussion and Conclusion}
\label{sec:conclusion}
  
We proposed SGE, an approach to increase state evaluation rates in constrained parallel search algorithm by separating successor generation and evaluation.
We showed SGE significantly increases the state evaluation rate of PUHF3 and OBAT, resulting in significantly improved overall performance and coverage. 
The batch successor insertion state bottleneck addressed by SGE (Section \ref{sec:bottleneck}) arises when implementing a parallel search algorithm which seeks to behave similarly to single-threaded GBFS with a standard eager evaluation policy, where states are evaluated immediately after they are generated and before being inserted in $\open$.
In lazy (deferred) evaluation \cite{RichterH09}, where states are not evaluated before insertion into $\open$ and are inserted into $\open$ based on their parent's $f$-value (and later evaluated when they are expanded), the batch successor insertion bottleneck would not apply. 
Search with lazy evaluation behaves quite differently than search with eager evaluation, and parallel satisficing search with lazy evaluation is an avenue for future work.

\section*{Acknowledgments}
This research was supported by JSPS KAKENHI Grant 20K11932 and JST SPRING Grant 250800000426.

\bibliography{references}

\clearpage
\onecolumn

\section*{Supplement}

\begin{figure}[H]
    \centering
    \begin{subfigure}[]{0.18\columnwidth}
      \includegraphics[width=\textwidth,trim={1cm 0.4cm 1cm 0.1cm},,clip]{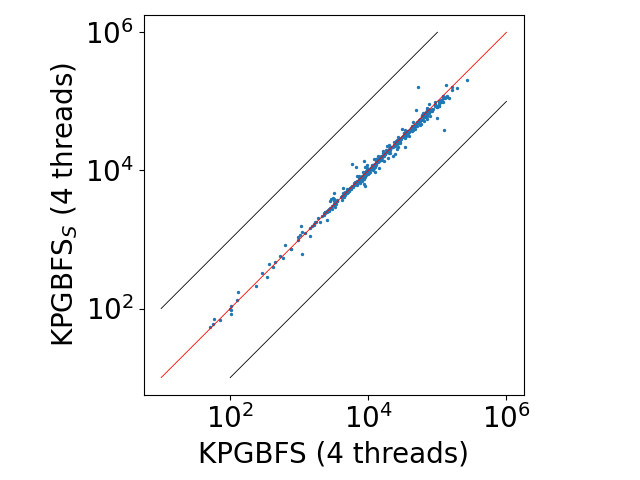}
      \caption{}
    \end{subfigure}
    \begin{subfigure}[]{0.18\columnwidth}
      \includegraphics[width=\textwidth,trim={1cm 0.4cm 1cm 0.1cm},,clip]{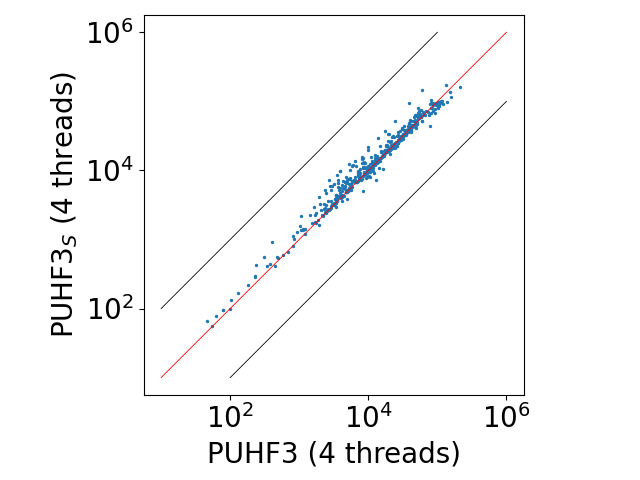}
      \caption{}
    \end{subfigure}
    \begin{subfigure}[]{0.18\columnwidth}
      \includegraphics[width=\textwidth,trim={1cm 0.4cm 1cm 0.1cm},,clip]{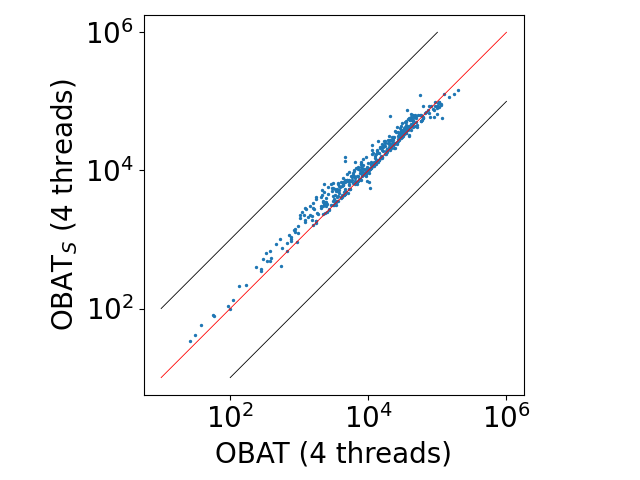}
      \caption{}
    \end{subfigure}
  
    \begin{subfigure}[]{0.18\columnwidth}
      \includegraphics[width=\textwidth,trim={1cm 0.4cm 1cm 0.1cm},,clip]{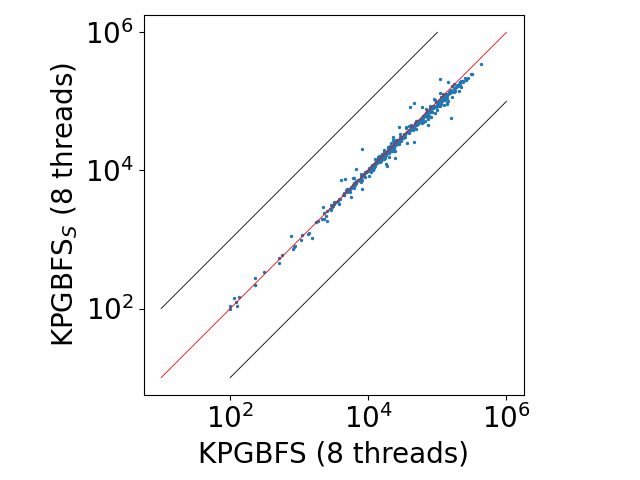}
      \caption{}
    \end{subfigure}
    \begin{subfigure}[]{0.18\columnwidth}
      \includegraphics[width=\textwidth,trim={1cm 0.4cm 1cm 0.1cm},,clip]{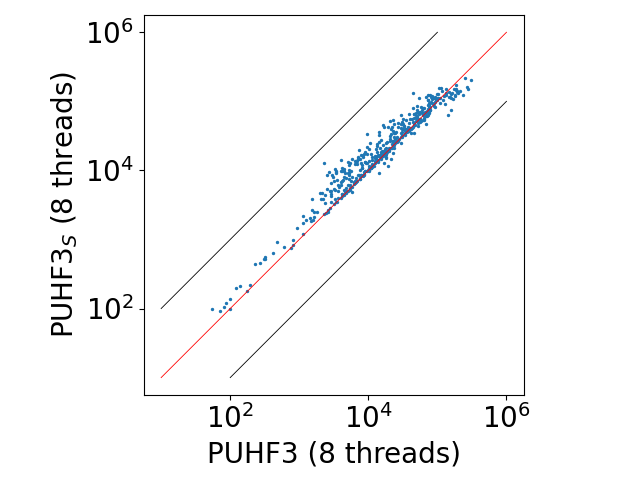}
      \caption{}
    \end{subfigure}
    \begin{subfigure}[]{0.18\columnwidth}
      \includegraphics[width=\textwidth,trim={1cm 0.4cm 1cm 0.1cm},,clip]{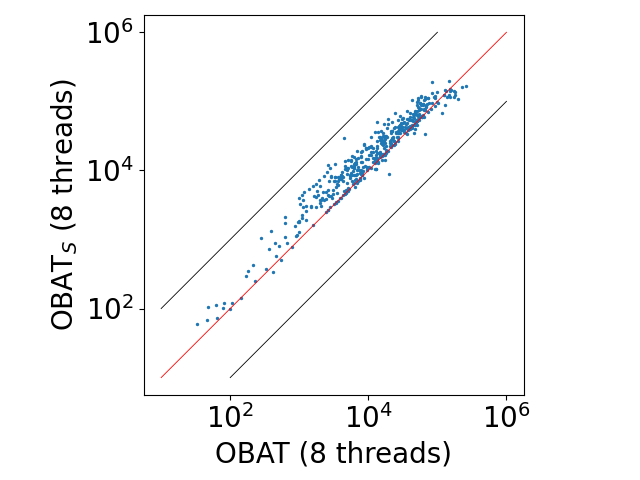}
      \caption{}
    \end{subfigure}
  
    \begin{subfigure}[]{0.18\columnwidth}
      \includegraphics[width=\textwidth,trim={1cm 0.4cm 1cm 0.1cm},,clip]{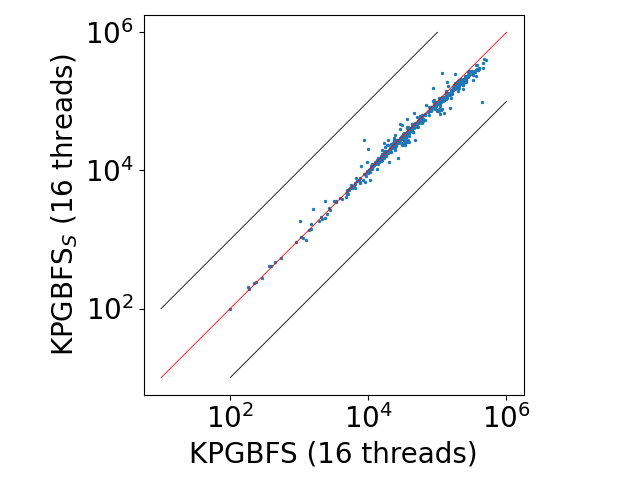}
      \caption{}
    \end{subfigure}
    \begin{subfigure}[]{0.18\columnwidth}
      \includegraphics[width=\textwidth,trim={1cm 0.4cm 1cm 0.1cm},,clip]{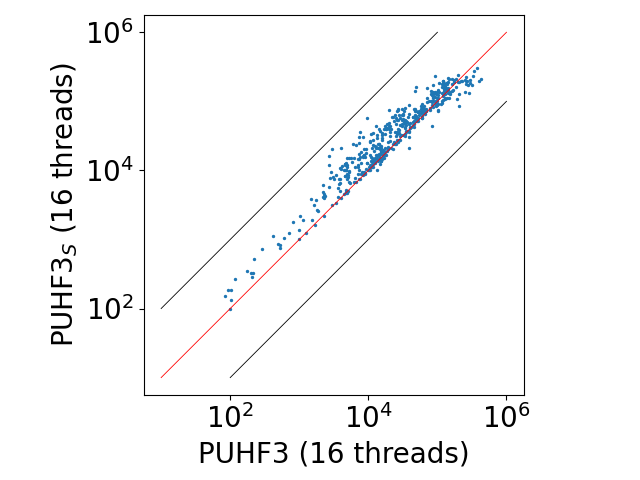}
      \caption{}
    \end{subfigure}
    \begin{subfigure}[]{0.18\columnwidth}
      \includegraphics[width=\textwidth,trim={1cm 0.4cm 1cm 0.1cm},,clip]{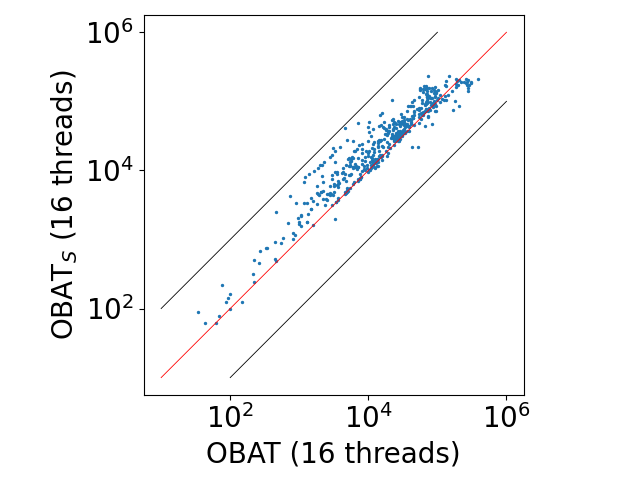}
      \caption{}
    \end{subfigure}
  
    \caption{ State evaluation rate comparison (states/second), Diagonal lines are $y=0.1x$, $y=x$, and $y=10x$}
    \label{sge-supp:fig:evaluation-rate-comparisons}
  
  \centering
  \end{figure}
  
\begin{figure}[H]
  \centering
  \begin{subfigure}[]{0.18\columnwidth}
    \includegraphics[width=\textwidth,trim={1cm 0.4cm 1cm 0.1cm},,clip]{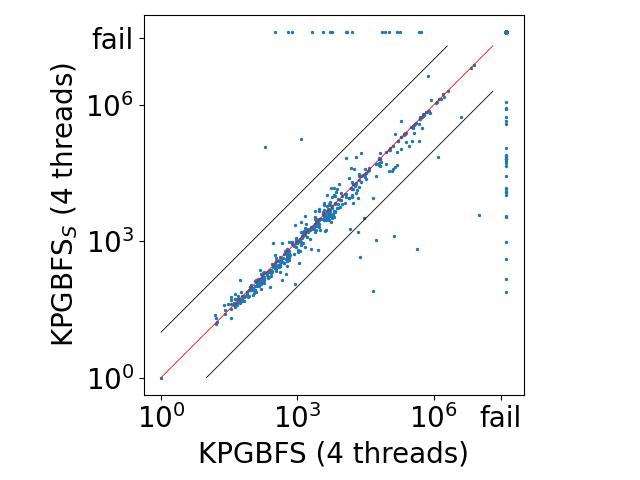}
    \caption{}
  \end{subfigure}
  \begin{subfigure}[]{0.18\columnwidth}
    \includegraphics[width=\textwidth,trim={1cm 0.4cm 1cm 0.1cm},,clip]{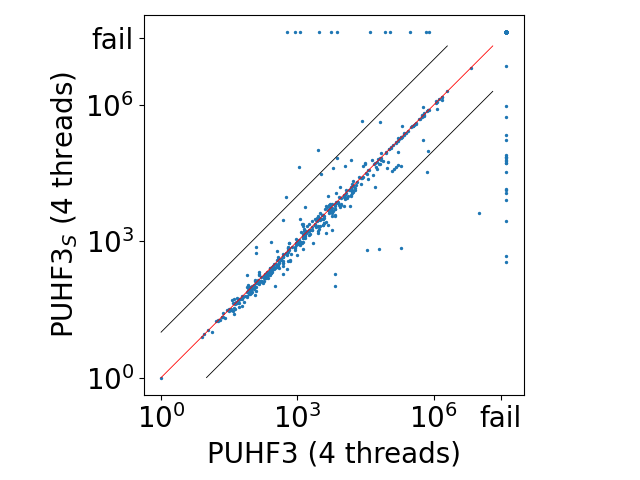}
    \caption{}
  \end{subfigure}
  \begin{subfigure}[]{0.18\columnwidth}
    \includegraphics[width=\textwidth,trim={1cm 0.4cm 1cm 0.1cm},,clip]{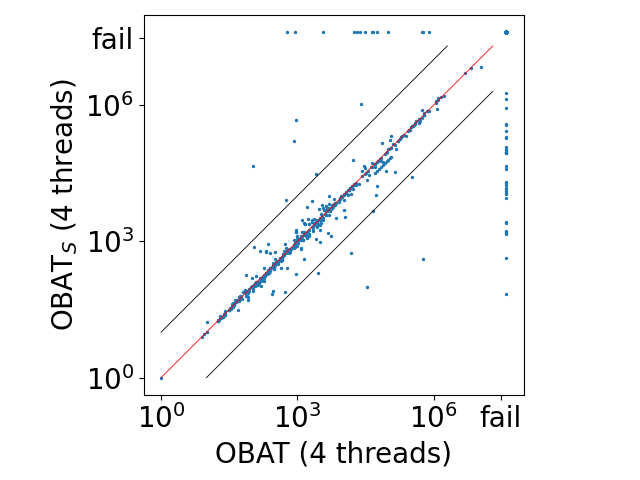}
    \caption{}
  \end{subfigure}

  \begin{subfigure}[]{0.18\columnwidth}
    \includegraphics[width=\textwidth,trim={1cm 0.4cm 1cm 0.1cm},,clip]{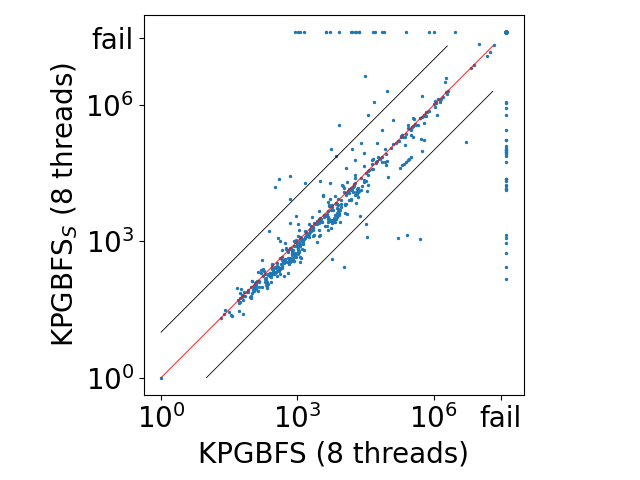}
    \caption{}
  \end{subfigure}
  \begin{subfigure}[]{0.18\columnwidth}
    \includegraphics[width=\textwidth,trim={1cm 0.4cm 1cm 0.1cm},,clip]{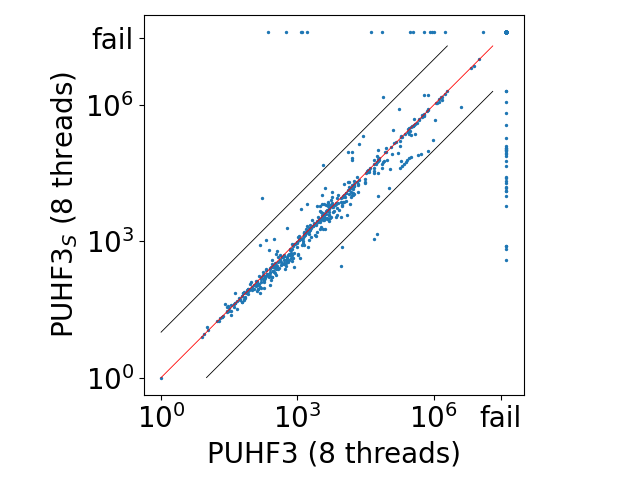}
    \caption{}
  \end{subfigure}
  \begin{subfigure}[]{0.18\columnwidth}
    \includegraphics[width=\textwidth,trim={1cm 0.4cm 1cm 0.1cm},,clip]{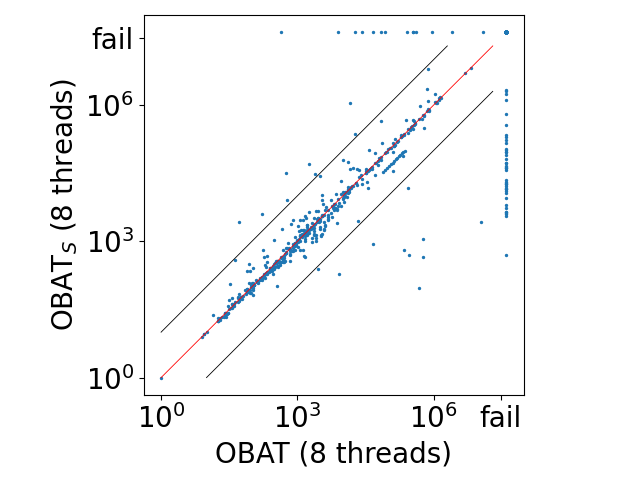}
    \caption{}
  \end{subfigure}

  \begin{subfigure}[]{0.18\columnwidth}
    \includegraphics[width=\textwidth,trim={1cm 0.4cm 1cm 0.1cm},,clip]{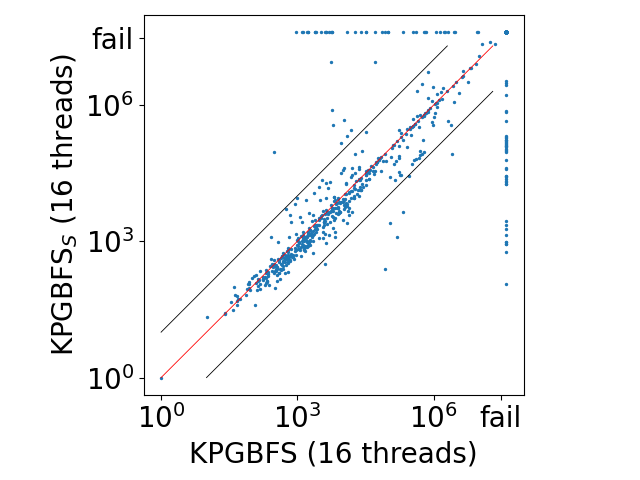}
    \caption{}
  \end{subfigure}
  \begin{subfigure}[]{0.18\columnwidth}
    \includegraphics[width=\textwidth,trim={1cm 0.4cm 1cm 0.1cm},,clip]{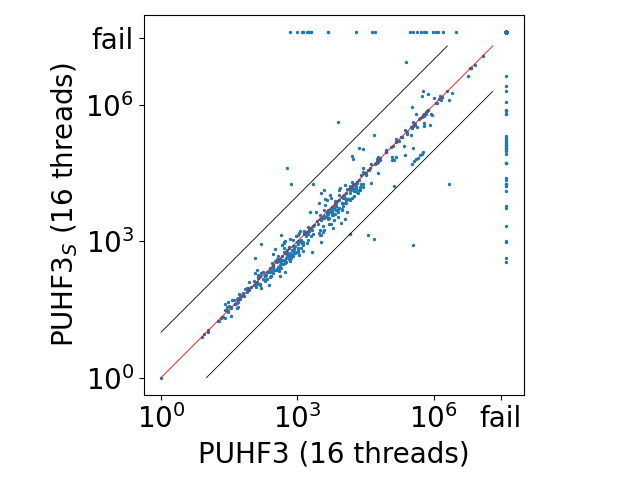}
    \caption{}
  \end{subfigure}
  \begin{subfigure}[]{0.18\columnwidth}
    \includegraphics[width=\textwidth,trim={1cm 0.4cm 1cm 0.1cm},,clip]{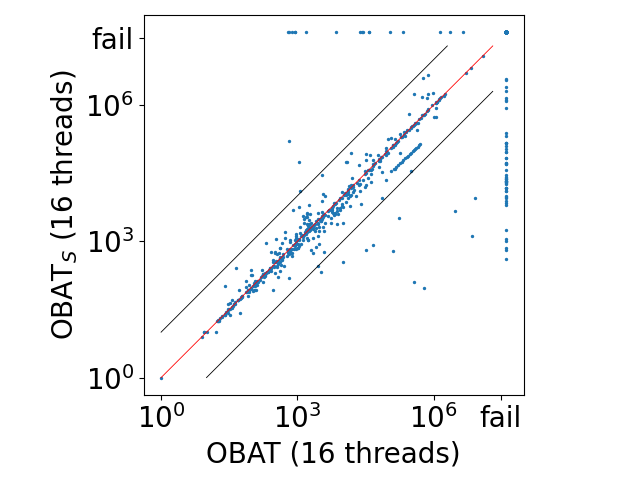}
    \caption{}
  \end{subfigure}

  \caption{ Number of states expanded, Diagonal lines are $y=0.1x$, $y=x$, and $y=10x$}
  \label{sge-supp:fig:expansions-comparisons}

\centering
\end{figure}

\begin{figure}[H]
  \centering
  \begin{subfigure}[]{0.18\columnwidth}
    \includegraphics[width=\textwidth,trim={1cm 0.4cm 1cm 0.1cm},,clip]{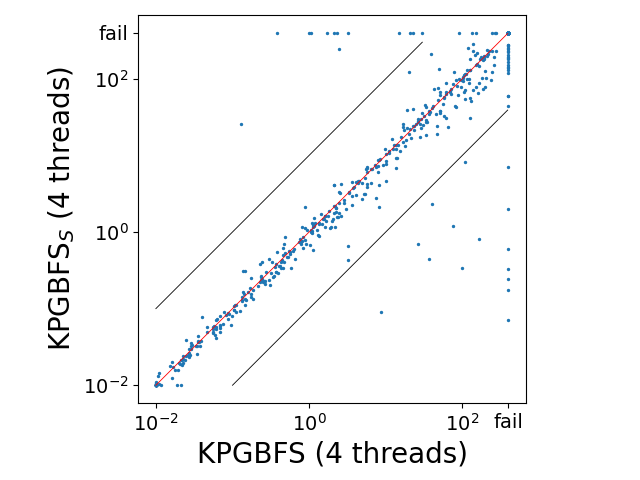}
    \caption{}
  \end{subfigure}
  \begin{subfigure}[]{0.18\columnwidth}
    \includegraphics[width=\textwidth,trim={1cm 0.4cm 1cm 0.1cm},,clip]{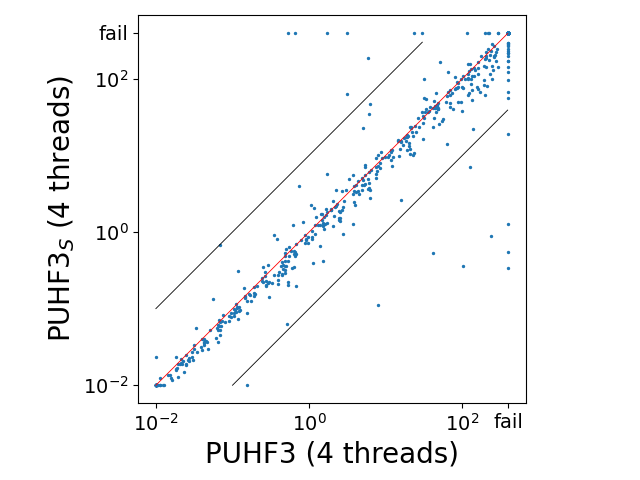}
    \caption{}
  \end{subfigure}
  \begin{subfigure}[]{0.18\columnwidth}
    \includegraphics[width=\textwidth,trim={1cm 0.4cm 1cm 0.1cm},,clip]{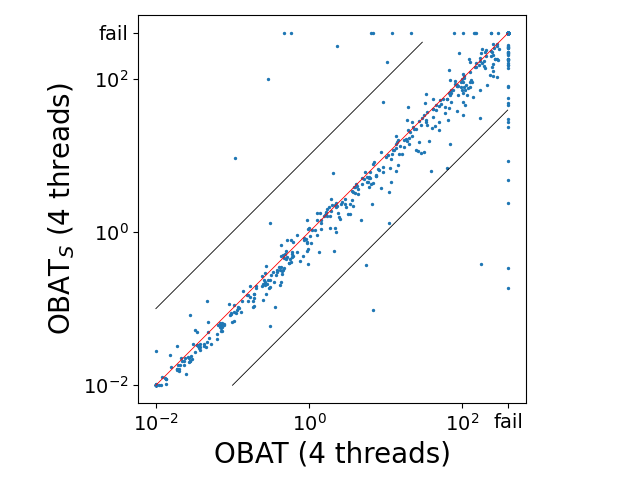}
    \caption{}
  \end{subfigure}

  \begin{subfigure}[]{0.18\columnwidth}
    \includegraphics[width=\textwidth,trim={1cm 0.4cm 1cm 0.1cm},,clip]{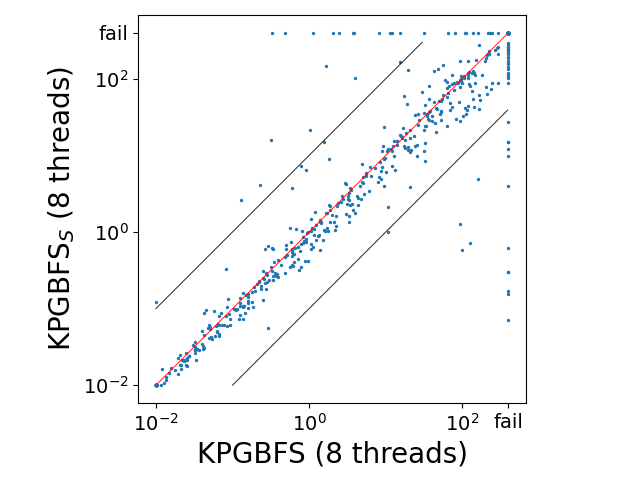}
    \caption{}
  \end{subfigure}
  \begin{subfigure}[]{0.18\columnwidth}
    \includegraphics[width=\textwidth,trim={1cm 0.4cm 1cm 0.1cm},,clip]{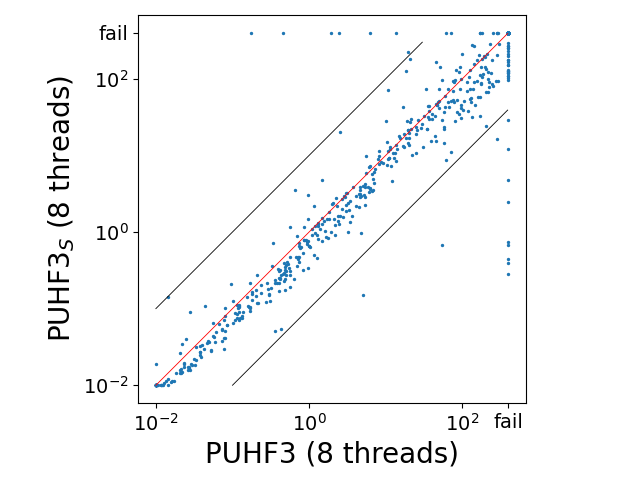}
    \caption{}
  \end{subfigure}
  \begin{subfigure}[]{0.18\columnwidth}
    \includegraphics[width=\textwidth,trim={1cm 0.4cm 1cm 0.1cm},,clip]{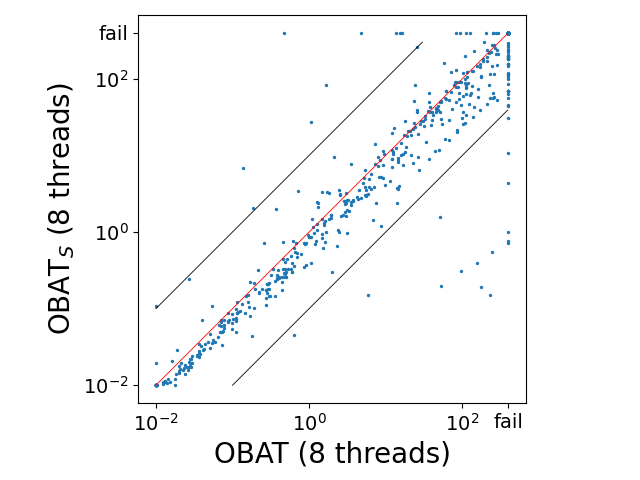}
    \caption{}
  \end{subfigure}

  \begin{subfigure}[]{0.18\columnwidth}
    \includegraphics[width=\textwidth,trim={1cm 0.4cm 1cm 0.1cm},,clip]{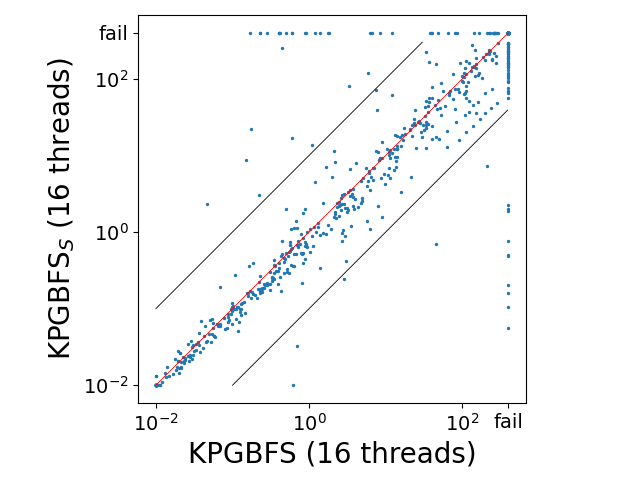}
    \caption{}
  \end{subfigure}
  \begin{subfigure}[]{0.18\columnwidth}
    \includegraphics[width=\textwidth,trim={1cm 0.4cm 1cm 0.1cm},,clip]{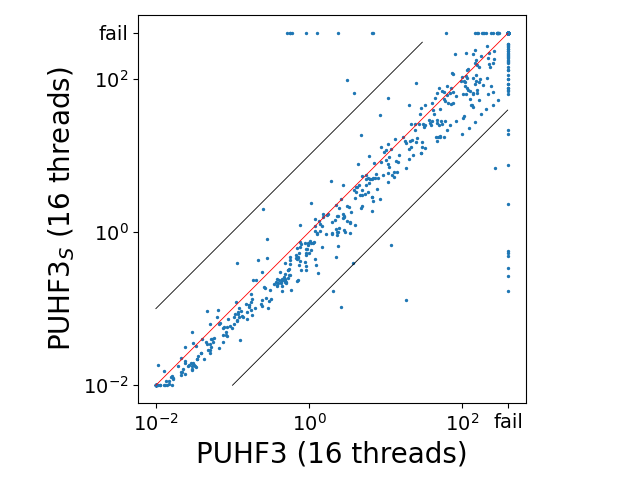}
    \caption{}
  \end{subfigure}
  \begin{subfigure}[]{0.18\columnwidth}
    \includegraphics[width=\textwidth,trim={1cm 0.4cm 1cm 0.1cm},,clip]{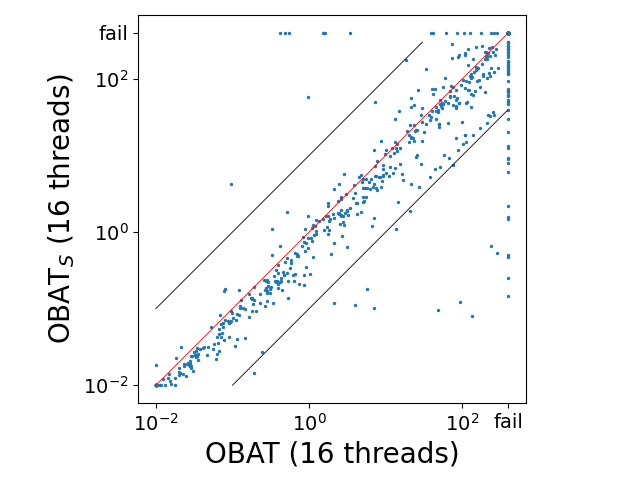}
    \caption{}
  \end{subfigure}
  \centering
  \caption{ Search time (seconds) ``fail''= out of time/memory, diagonal lines are $y=0.1x$, $y=x$, and $y=10x$}
  \label{sge-supp:fig:search-time-comparisons}
  \centering
\end{figure}
\end{document}